\documentclass[a4paper,twoside]{article}

\usepackage{epsfig}
\usepackage{subcaption}
\usepackage{calc}
\usepackage{amssymb}
\usepackage{amstext}
\usepackage{amsmath}
\usepackage{amsthm}
\usepackage{multicol}
\usepackage{pslatex}
\usepackage{apalike}
\usepackage[bottom]{footmisc}
\usepackage{SCITEPRESS}     

\begin{document}

\title{Evaluating Deep Learning Assisted Automated Aquaculture Net Pens Inspection Using ROV}

\author{\authorname{Waseem Akram\sup{1}, Muhayyuddin Ahmed\sup{1}, Lakmal Seneviratne\sup{1} and Irfan Hussain\sup{1}}
\affiliation{\sup{1}Khalifa University Center for Autonomous Robotic Systems (KUCARS), Khalifa University, United Arab Emirates.}
\email{\{waseem.akram, muhayyuddin.ahmed,lakmal.seneviratne,irfan.hussain\}@ku.ac.ae}
}

\keywords{Aquaculture, Net defect detection, Deep learning, Marine vehicle}

\abstract{
In marine aquaculture, inspecting sea cages is an essential activity for managing both the facilities' environmental impact and the quality of the fish development process. Fish escape from fish farms into the open sea due to net damage, which can result in significant financial losses and compromise the nearby marine ecosystem. The traditional inspection system in use relies on visual inspection by expert divers or ROVs, which is not only laborious, time-consuming, and inaccurate but also largely dependent on the level of knowledge of the operator and has a poor degree of verifiability. This article presents a robotic-based automatic net defect detection system for aquaculture net pens oriented to on-ROV processing and real-time detection. The proposed system takes a video stream from an onboard camera of the ROV, employs a deep learning detector, and segments the defective part of the image from the background under different underwater conditions. The system was first tested using a set of collected images for comparison with the state-of-the-art approaches and then using the ROV inspection sequences to evaluate its effectiveness in real-world scenarios. Results show that our approach presents high levels of accuracy even for adverse scenarios and is adequate for real-time processing on embedded platforms.}

\onecolumn \maketitle \normalsize \setcounter{footnote}{0} \vfill

\section{\uppercase{Introduction}}
\label{sec:introduction}

\noindent Marine aquaculture has become an essential part of meeting the growing demand for high-quality protein while also preserving the ocean environment \cite{gui2019research}. There are various types of marine aquaculture facilities, such as deep-sea cage farming, raft farming, deep-sea platform farming, and net enclosure farming \cite{yan2018research}. Advancements in engineering and construction have significantly improved the wave resistance of aquaculture facilities like cages and net enclosures \cite{zhou2018current}. However, the netting used in these facilities is a crucial component that is easily damaged and difficult to detect, leading to the escape of cultured fish and causing substantial economic losses every year. The detection of net damage is currently impeding the development of marine aquaculture facility \cite{wei2020intelligent}.

\begin{figure}[t]
\centering
\subfloat[ROV.  \label{fig:B2}]{\includegraphics[width=0.5\columnwidth, height=0.3\columnwidth]{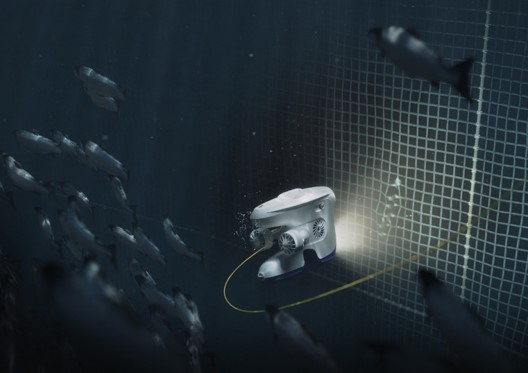}}
\subfloat[Net hole. \label{fig:B1}]{\includegraphics[width=0.5\columnwidth,height=0.3\columnwidth]{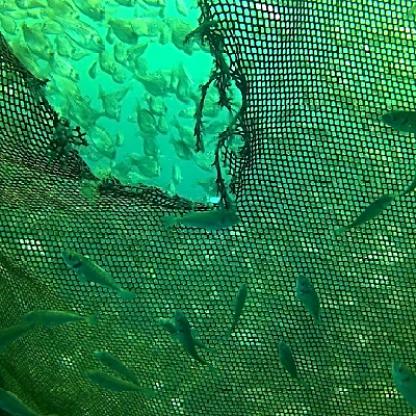}}
\vspace{5pt}\hspace{0.2cm}
\subfloat[Plastic. \label{fig:B3}]{\includegraphics[width=0.5\columnwidth, height=0.3\columnwidth]{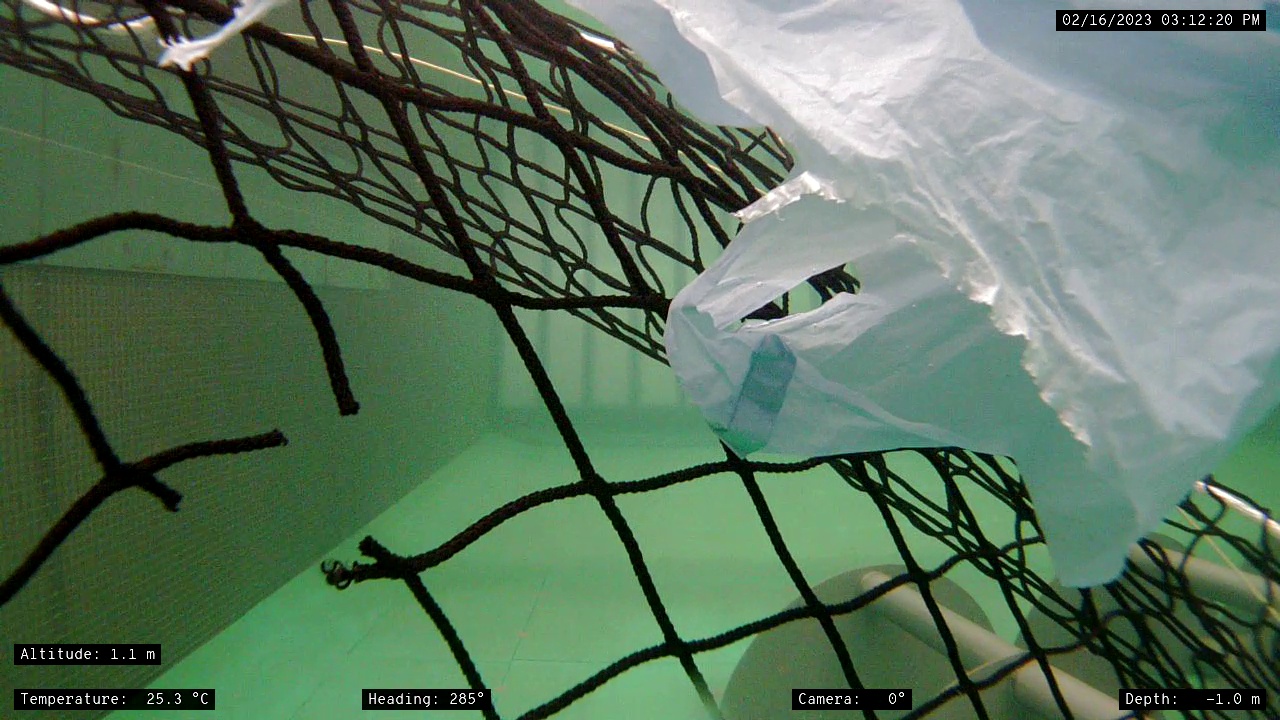}}
\subfloat[Vegetation.  \label{fig:B4}]{\includegraphics[width=0.5\columnwidth, height=0.3\columnwidth]{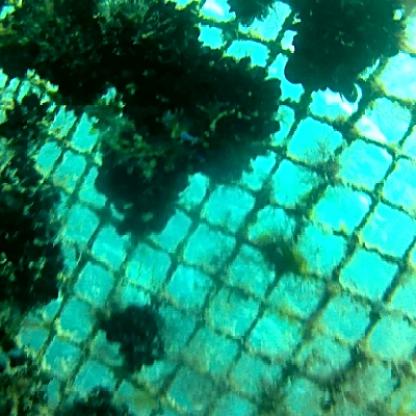}}\vspace{5pt}\hspace{0.2cm}
\caption{Automatic aquaculture inspection using ROV. (a) shows an ROV during aquaculture net pens inspection, (b) shows an example of the hole on the net, (c) shows an example of plastic entangled on the net, and (d) shows an example of vegetation attached to the net.}
\label{fig:ins}
\end{figure}

\noindent At present, the standard practice for identifying damage in sea cages involves human visual inspection, which can be carried out directly by professional divers or remotely by observing video feeds captured with Remotely Operated Vehicles (ROVs) \cite{wei2020intelligent} as shown in Figure \ref{fig:ins}. However, relying on professional divers for inspections can be expensive, time-consuming, and poses safety risks. Furthermore, it provides low coverage, verifiability, and repeatability \cite{akram2022visual}. In contrast, analyzing images captured by ROVs equipped with different types of cameras eliminates safety risks and offers the potential to implement computer-assisted damage detection procedures through computer vision techniques. Recent developments have shown the use of deep learning approaches \cite{sun2020deep,yang2021deep} such as Mask R-RCNN \cite{maskrcnn}, Fast-RCNN \cite{fastrcnn}, YOLO \cite{yolov5}, and SSD \cite{ssd} for automatic aquaculture net damage detection. 

\noindent Various sensor modalities, such as cameras or acoustic-based sensors, are used for automatic inspection operations of net cage structure integrity. Computer vision-based approach, such as \cite{zhang2022netting} proposed by Zhang et al. uses Mask-RCNN \cite{maskrcnn} for net hole detection problems in a laboratory setup. The experimental results demonstrated an Average Precision score of 94.48\%. Similarly, Liao et al. \cite{liao2022research} proposed a MobileNet-SSD network model for hole detection in open-sea fish cages. In this work, net hole detection is performed by fusing MobileNet with the SSD network model. The results showed an average precision score of 88.5\%. Tao et al. \cite{tao2018omnidirectional}, applied deep learning to detect net holes using the YOLOv1 (You Only Look Once) algorithm on images captured under controlled lab conditions. In contrast, Madshaven et al. \cite{madshaven2022hole}, employed a combination of classical computer vision and image processing techniques for tracking, alongside neural networks for segmenting the net structure and classifying scene content, including the detection of irregularities caused by fish or seaweed. Qiu et al. \cite{qiu2020fishing}, utilized image-enhancing methods for net structure analysis and marine growth segmentation.

\noindent In previous literature on aquaculture inspection and monitoring perspective, most of the studies have focused on the hole detection problem. However, apart from the net hole, other serious problems can damage the net structure. For example, vegetation can grow and become attached to the net pens and cause fouling, which can reduce water flow and oxygenation. They can also damage the net which increases the risk of escape for the fish. In addition, plastic waste can entangle and harm fish, reduce water flow and oxygenation, and leach harmful chemicals into the water. To mitigate the negative effects of net holes, vegetation, and plastic waste on aquaculture net pens, it is important to implement proper net defect detection methods to perform regular inspection and monitoring activity to prevent the accumulation and damage of these net abnormalities.

\subsection{Contributions}
\noindent This paper introduces deep learning coupled with the ROV method that focuses on detecting irregularities in fish-cage nets, which is a critical task within the full inspection process. The method proposed in this paper employs a deep learning-based detector to identify areas where potential net defects such as plants, holes, and plastic exist in the net structure. We have evaluated different variants of the YOLO deep learning model to identify the net defects in real time using the ROV. The proposed approach leverages traditional computer vision and image processing methods and works under realistic lighting conditions. The method has been tested at a lab pool setup for fish nets.

\section{Proposed method}
\noindent Our proposed method for identifying net damage involves utilizing both image and video processing techniques in a cohesive workflow. The approach aims to identify and track net defects such as vegetation, holes, and plastic in the aqua-net. To achieve this, a remotely operated vehicle (ROV) named Blueye Pro ROV X, equipped with an HD camera is utilized to conduct a controlled inspection campaign in the Marine Pool at Khalifa University, UAE. The camera remains at a fixed distance from the area of interest throughout the inspection, with the option to adjust the distance as needed. As shown in Figure, the net is 10x10m long mesh placed in the pool where vegetation and plastic are attached to the net. In addition, there are net holes at different location on the net surface. The net is spread around the side of the pool, and the ROV is allowed to inspect the net at constant speed to record the experimental data. 

\begin{figure}[t]
\centering
\includegraphics[width=1\columnwidth]{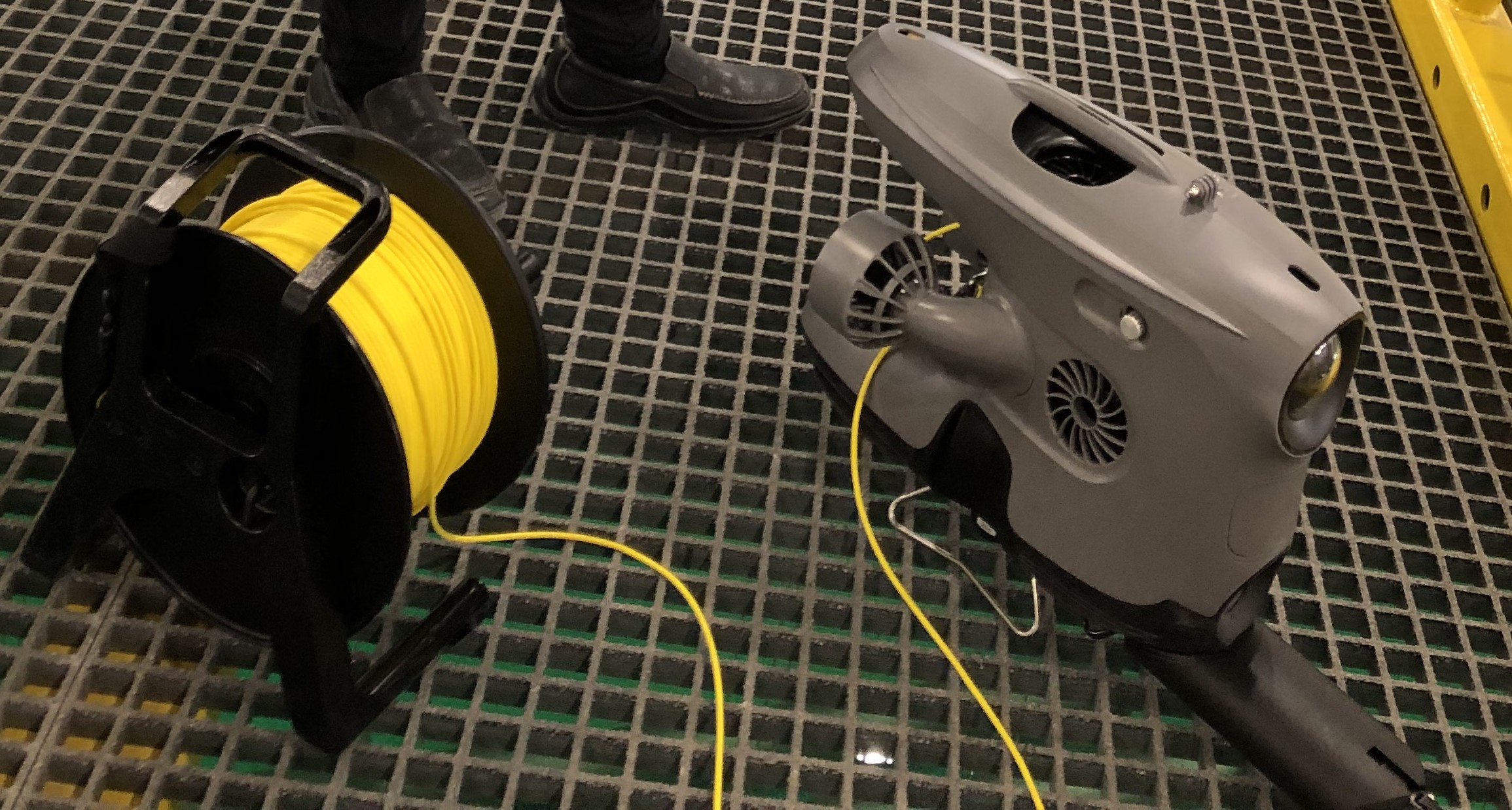}
\caption{The Blueye Prov ROV x deployed for aqua-net inspection.}
\label{fig:rovx}
\end{figure}

\noindent The ROV as shown in Figure \ref{fig:rovx}, has six thrusters and is equipped with Doppler Velocity Log (DVL), Inertial Measurement Unit (IMU), and camera sensor. The ROV can operate in saltwater, brackish water, or freshwater for up to 2 hours on a charge and can descend to a depth of 300 meters when tethered by a cable measuring 400 meters in length. In addition, a forward-facing camera capable of 25-30 frames per second (fps) in full HD has been placed for use in making high-quality video streams of the surrounding area. The captured video feed is then used as input for the processing flow.

\noindent The video that was captured has dimensions of 1920 x 1080 and a frame rate of 30 frames per second. To extract image sequences of a total of 510, we captured one frame every five seconds from the video. For training, we utilized 90\% images from the dataset, while the remainder was allocated for testing.

\noindent To validate and show the effectiveness of the method, the ROV is subjected to perform the real-time detection of the aqua net defects. To achieve this, we deploy the ROV in the pool. The initial position of the ROV is at a constant distance parallel to the net. The ROV communicates with the surface computer via a tether line for video transmission. Moreover, the ROV moment is manually controlled through a remote controller. The online camera recording was then used on the surface computer and given to the detector module. The detector module was allowed to display the detection along with a bounding box.

\subsection{Detection Approaches}
\noindent Deep learning models have proven to be highly effective in detecting and precisely locating the affected region within an input image. Numerous cutting-edge models have been introduced in the research literature to address this task. In our study, we conducted tests and evaluations on several widely employed detection models, namely YOLOv4 \cite{yolov4}, YOLOv5 \cite{yolov5}, YOLOv7 \cite{yolov7}, and YOLOv8 \cite{yolov8}, to identify and localize vegetation, plastic, and holes using the Aqua-Net detection method.

\noindent YOLOv4 represents an enhanced iteration of YOLOv3, surpassing it in terms of mean average precision (mAP) and frames per second (FPS). It is recognized as a one-stage object detection approach, which comprises three key components: backbone, neck, and head. The backbone component, referred to as CSPDarknet53, is a pre-trained Convolutional Neural Network (CNN) with 53 CNN layers. Its principal objective is to extract features and generate feature maps from the input images. The neck component incorporates spatial pyramid pooling (SPP) and a path aggregation network (PAN), linking the backbone to the head. Lastly, the head component processes the aggregated features and makes predictions regarding bounding boxes and classification.

\noindent The architecture of YOLOv5 is based on a backbone network, typically a CNN, which extracts features from the input image. These features are then passed through several additional layers, including convolutional, upsampling, and fusion layers, to generate high-resolution feature maps. The model further utilizes anchor boxes, which are pre-defined boxes of various sizes and aspect ratios, to predict bounding boxes for objects. YOLOv5 predicts the coordinates of bounding boxes relative to the grid cells and refines them with respect to anchor boxes. To train the YOLOv5 model, a large labeled dataset is required, along with bounding box annotations for the objects of interest. The model is trained using techniques like backpropagation and gradient descent to optimize the network parameters and improve object detection performance. YOLOv5 offers different model sizes (e.g., YOLOv5s, YOLOv5m, YOLOv5l, and YOLOv5x) that vary in terms of depth and computational complexity. These different variants provide a trade-off between speed and accuracy, allowing users to choose the model that suits their specific requirements.

\noindent Yolov7s represents a compact variant of YOLOv7, which is an improved version of YOLOv6 in terms of mean average precision (mAP), detection speed, and inference performance. YOLOv7 introduces an extension of the Efficient Layer Aggregation Network (ELAN) called extended ELAN (EELAN). The network incorporates various fundamental techniques, such as expand, shuffle, and merge, to enhance its learning capability without disrupting the gradient flow. These enhancements are aimed at improving the network's performance while maintaining its efficiency and preserving the flow of gradients. In addition, YOLOv7 also emphasizes the utilization of methods such as trainable "bag-of-freebies" and optimization modules. The model incorporates different computational blocks to extract more distinct features. These feature maps from each computational block are grouped into sets of a specific size, denoted as "s," and subsequently concatenated. Finally, the cardinality is merged by shuffling the group feature map, completing the overall process.

\noindent The YOLOv8n model, as the nano edition of the YOLOv8 family, stands out for its compact size, exceptional speed, and improved detection capabilities. It is particularly well-suited for object detection and classification tasks. Moreover, when combined with instance segmentation and object tracking techniques, it proves to be at the forefront of the State-of-the-Art (SOTA), delivering advanced performance in various computer vision applications. The model is characterized as anchor-free, indicating that instead of estimating the deviation from a predefined anchor box, it directly predicts the center point of an object in an image. In YOLOv8, image augmentation is performed at each epoch to enhance the training process. The technique employed is mosaic augmentation, where four images are stitched together to create a mosaic, thereby encouraging the model to learn from new spatial arrangements. One notable change in the YOLOv8 architecture is the replacement of the c3 module with the c2f module. In the c2f module, the outputs from Bottleneck layers are concatenated, whereas in the previous c3 module, only the output from the last Bottleneck layer was utilized. Additionally, the first set of 6x6 convolutional layers in the Backbone module is substituted with a 3x3 convolutional block, bringing about a structural modification in YOLOv8.

\section{Implementation and Experimental setup}
\noindent In order to perform the aquaculture net defect detection including vegetation, plastic, and net holes, the system is tested in real-time using ROV. The experimental environment is developed on a core i9-10940@3.30 GHz processor, 128GB RAM, and a single NVIDIA Quadro RTX 6000 GPU equipped with a CUDA toolkit. Python version 3.8.10 was utilized, alongside the PyTorch 1.13.1 cuda 11.7 framework. Additionally, the model underwent training for 300 epochs, with each epoch consisting of 512 iterations. The collected images having a resolution of 1920 x 1080 pixels are annotated by the LabelImg annotation tool, and the TXT annotation file in YOLO format is created, then the ratio of training set and testing set is set to 9:1.

\section{Results and Discussions}
\begin{figure*}[t]
\centering
\includegraphics[width=1\linewidth]{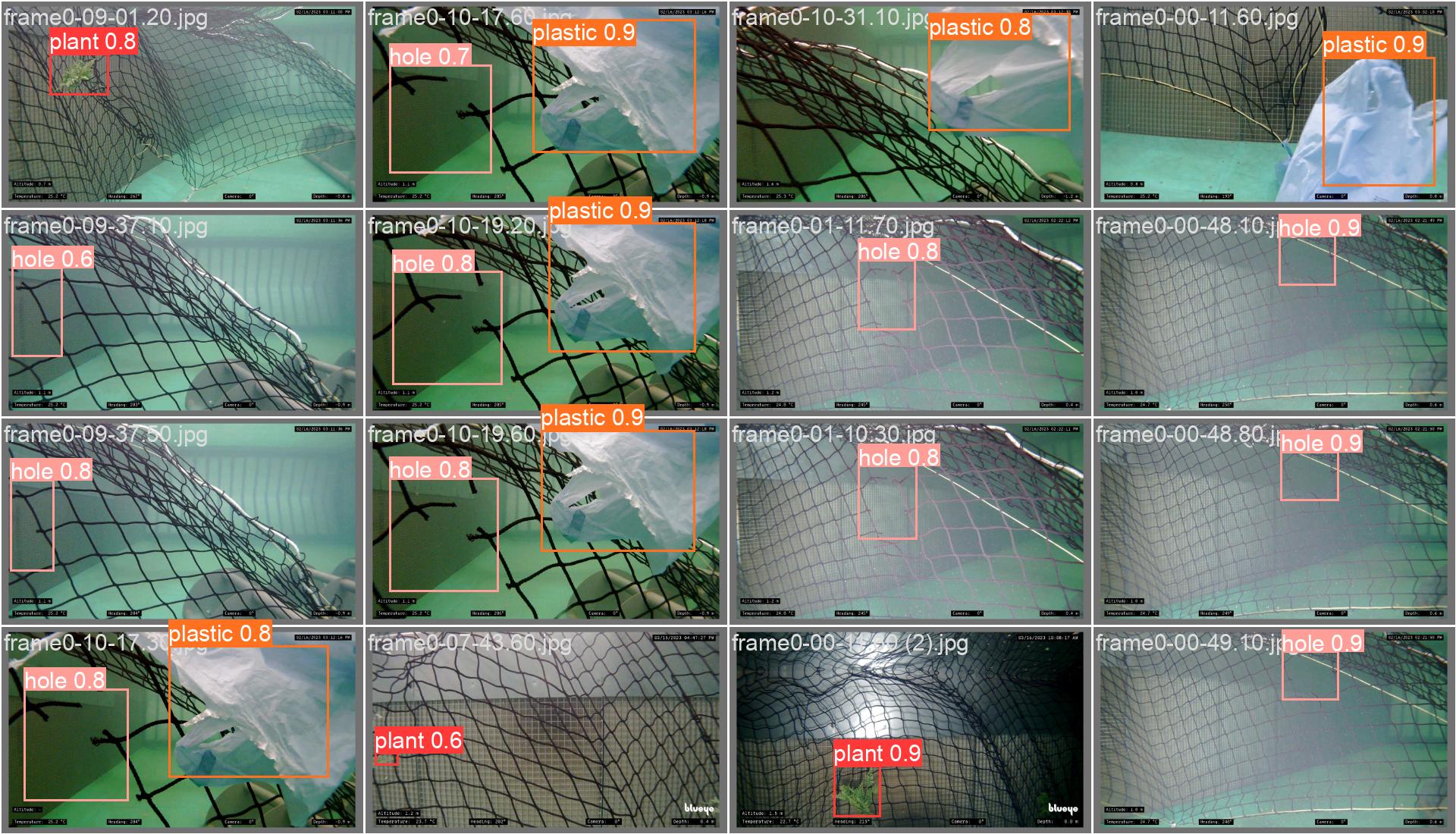}
\caption{Example of successful detection of aqua-net defects using YOLOv5.}
\label{fig:val}
\end{figure*}
\noindent The best-trained model was analyzed and embedded on the ROV for real-time detection which helps the user to timely detect the aqua-net defects within aquaculture. First, the experimental results of the YOLOv4, YOLOv5, YOLOv7, and YOLOv8 are shown in Table \ref{tab:rst}. According to Table \ref{tab:rst}, YOLOv5 has an advantage over other YOLO model variants in terms of mAP metrics. YOLOv4 showed a better precision score than the other models. In terms of Recall, YOLOv4 has achieved a 100\% precision score. Similarly, a higher F1 score was achieved by all models. It is noted that the YOLOV7 has less detection performance compared to other YOLO variants.


\begin{table}[t]
\footnotesize
    \centering
    \caption{Performance of YOLOV detectors.}
    \begin{tabular}{c c c c c}
         \hline
         Detector & mAP & Precision & Recall & F1 score \\\hline
         YOLOv4 & 0.9901& 0.9998& 0.9980& 0.9990\\
         YOLOv5 &0.9950 & 0.9895 & 0.9901 &0.9922 \\
         YOLOv7 &0.9656 &0.9698 &0.9006 &0.9290 \\
         YOLOv8 &0.9946 &0.9766 &0.9998 & 0.9881\\
         \\\hline

    \end{tabular}
    \label{tab:rst}
\end{table}

\noindent The aqua-net detection visualization results are shown in Figure \ref{fig:val}. Here, the analysis was performed to view the detected aqua-net defects i.e., plant, holes, and plastic after training the YOLOv5 model. The model was deployed on the unseen real-time images in a pool set up to check the feasibility of the model after training on the collected custom dataset. The threshold value was set to 0.3, which means if the confidence score is greater or equal to 30\% then the model categorizes it into the relevant class. In some cases, the percentage of defections is a little lower. The performance variation is due to the presence of different inputs in real time. Moreover, the model showed the ability to detect the net defect of different sizes as can be noted in Figure \ref{fig:val}.

\begin{figure}[t]
\centering
\includegraphics[width=1\columnwidth]{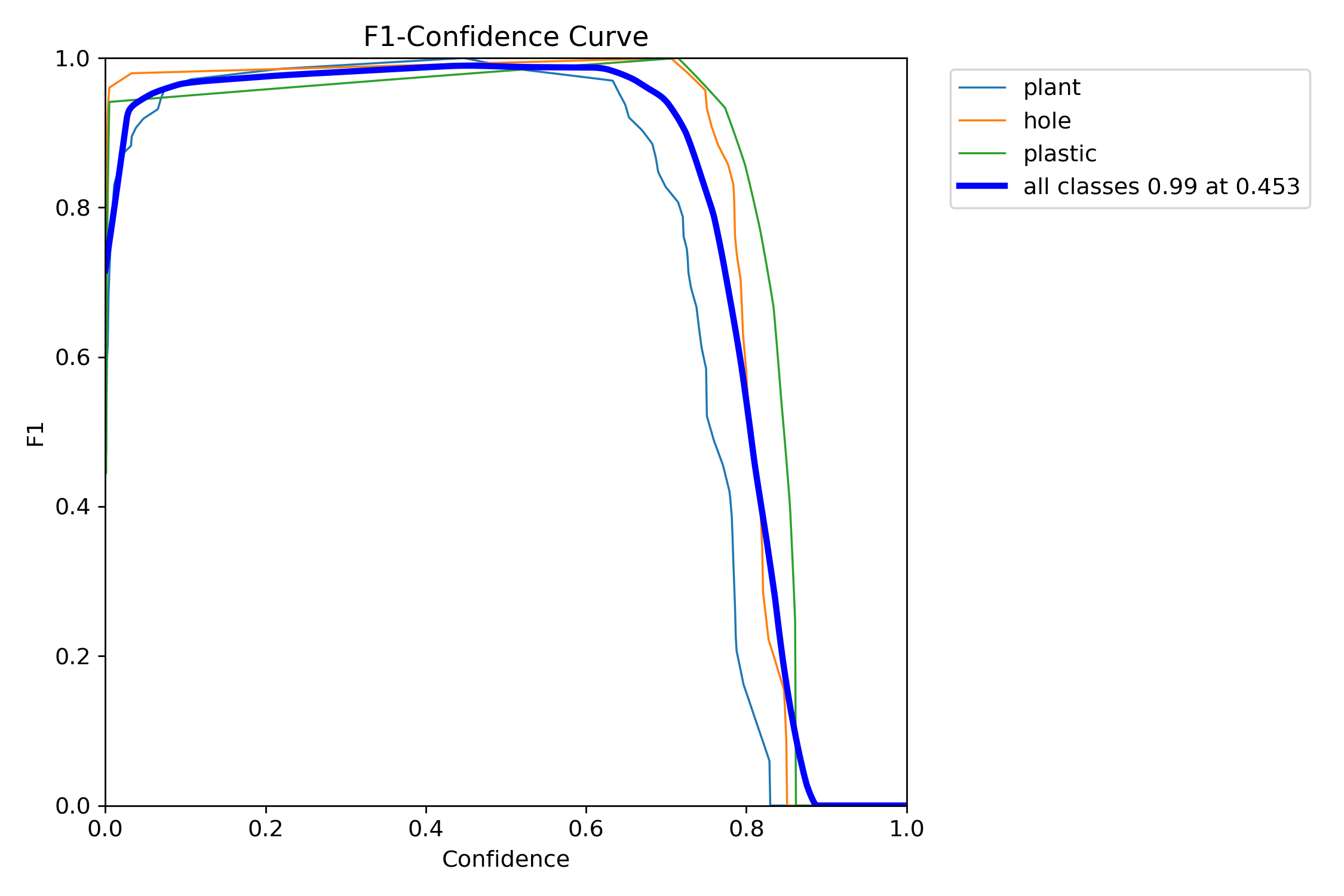}
\caption{Performance evaluation of the model through F1 curve.}
\label{fig:fc}
\end{figure}

\begin{figure}[t]
\centering
\includegraphics[width=1\columnwidth]{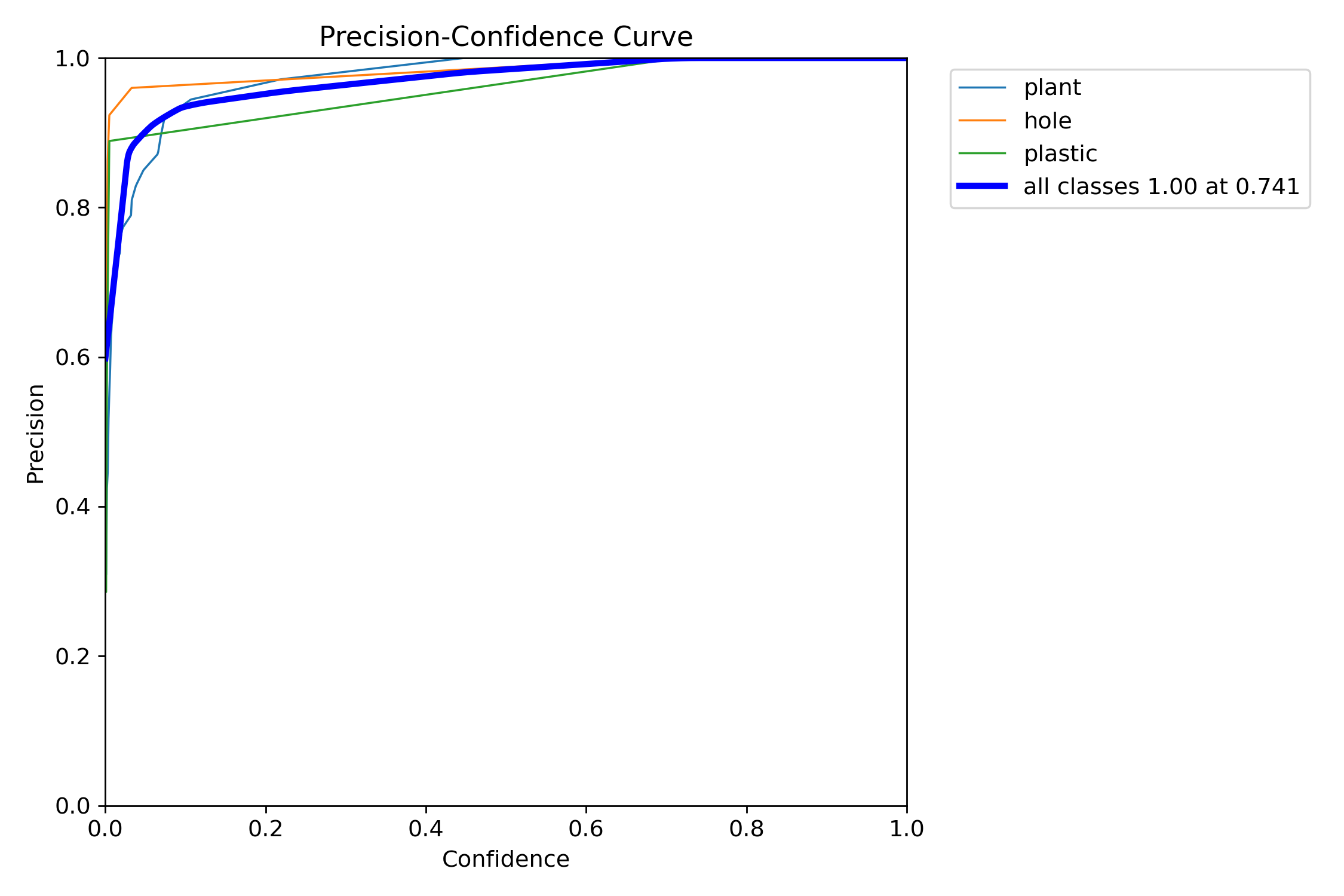}
\caption{Performance evaluation of the model through Precision curve curve.}
\label{fig:pc}
\end{figure}

\begin{figure}[t]
\centering
\includegraphics[width=1\columnwidth]{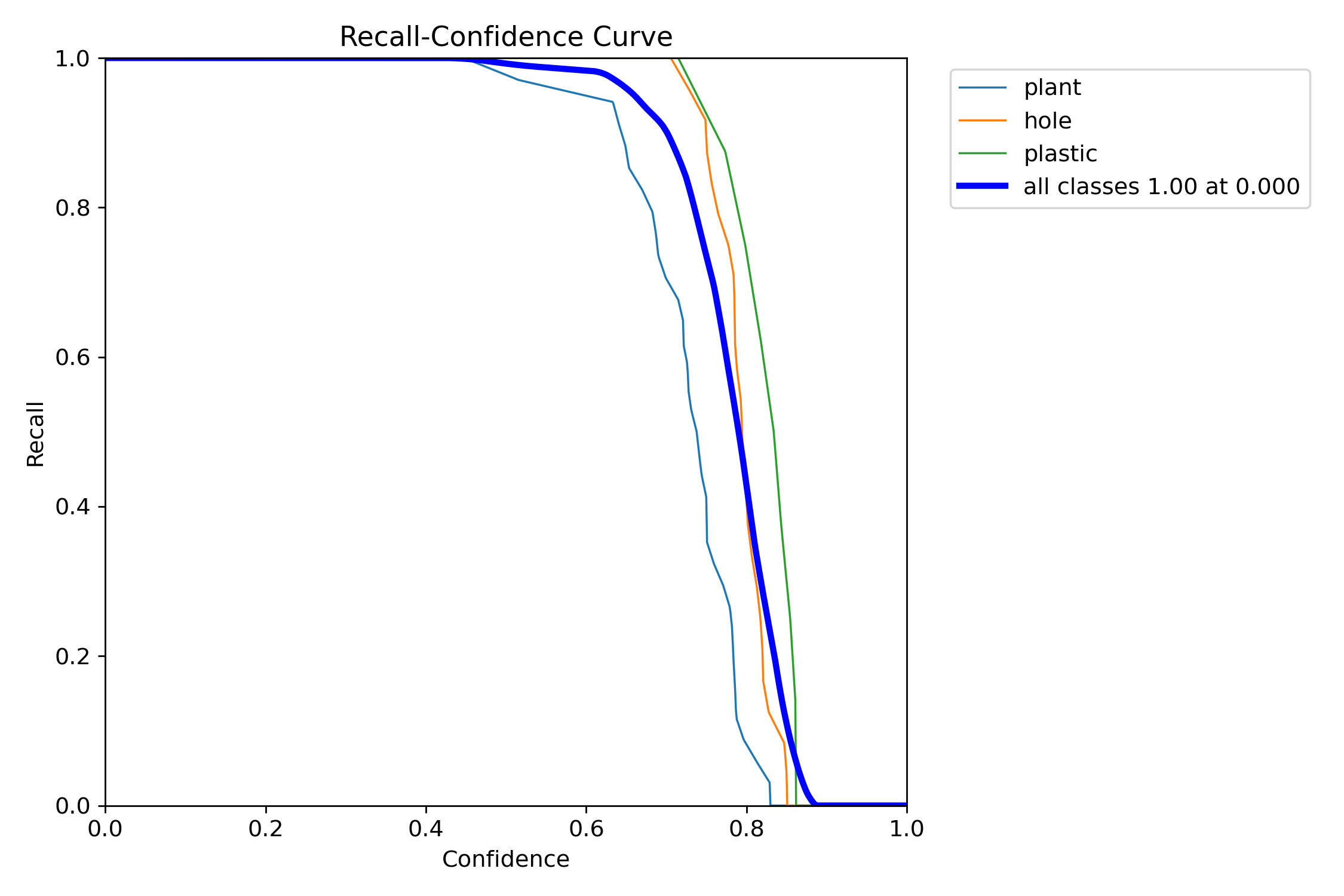}
\caption{Performance evaluation of the model through Recall curve.}
\label{fig:rc}
\end{figure}

\noindent Moreover, the performance of the detection yolov5 model is shown in Figures \ref{fig:fc}, \ref{fig:pc}, and \ref{fig:rc} in terms of F1 curve, Precision curve, and Recall curve. In Figure \ref{fig:fc}, among all three classes i.e., plant, hole, and plastic, the hole class achieved higher at the start, while the plastic class has the advantage over the others after training completion. However, the plant class achieved fewer scores than the others. Furthermore, it is seen that the F1 scores of all classes are non significantly different from each other. The Recall curve as shown in Figure \ref{fig:rc} demonstrates better detection performance for both plastic and hole classes. Similarly, the precision curve as shown in Figure \ref{fig:pr} also demonstrated better detection performance for all three classes i.e., plant, hole, and plastic aqua-net defects. It can also be noted that as the confidence level reaches 0.741, the Precision of 100\% is achieved by the adopted model. The obtained results indicate the effectiveness of the YOLOv5 detector for the aqua-net defect detection tasks.

\noindent To further evaluate the detection performance of our current research, we took into account the studies conducted by Zhang et al. in \cite{zhang2022netting} and Liao et al. in \cite{liao2022research}. In the domain of computer vision, Zhang et al. introduced a method that utilizes Mask-RCNN \cite{maskrcnn} to address the net hole detection problem within a controlled laboratory environment. Their experimental findings exhibited an Average Precision score of 94.48\%. Similarly, Liao et al. proposed a network model called MobileNet-SSD for detecting holes in fish cages located in open-sea environments. In their work, net hole detection is achieved by combining MobileNet with the SSD network model. The results demonstrated an average precision score of 88.5\%. Our study employed YOLOv4, YOLOv5, YOLOv7, and YOLOv8 models to detect aqua-net defects, and we achieved a 99\% performance across all metrics, including mAP, precision, recall, and F1 score. These results are presented in Table \ref{tab:rst}.

\section{\uppercase{Conclusions}}
\label{sec:conclusion}
\noindent In this research paper, we introduce the utilization of YOLO-aqua-net models for detecting plant, holes, and plastic within an aquaculture environment. The study involved the use of datasets acquired from a laboratory setup, where an ROV was employed to gather, test, and validate the proposed approach. Deep learning models based on YOLOV architecture were trained and evaluated to detect various defects in aqua-nets. The results demonstrate that all variations of YOLO models exhibit better real-time detection of aqua-net defects, making them suitable for real-time inspection tasks in aquaculture net pens.

\vfill
\section*{\uppercase{Acknowledgements}}

\bibliographystyle{apalike}
{\small
\bibliography{Example}}

\end{document}